\documentclass[letterpaper]{article} % DO NOT CHANGE THIS
\usepackage{aaai2026}  % DO NOT CHANGE THIS
\usepackage{times}  % DO NOT CHANGE THIS
\usepackage{helvet}  % DO NOT CHANGE THIS
\usepackage{courier}  % DO NOT CHANGE THIS
\usepackage[hyphens]{url}  % DO NOT CHANGE THIS
\usepackage{graphicx} % DO NOT CHANGE THIS
\usepackage{natbib}  % DO NOT CHANGE THIS AND DO NOT ADD ANY OPTIONS TO IT
\usepackage{caption} % DO NOT CHANGE THIS AND DO NOT ADD ANY OPTIONS TO IT
\usepackage{algorithm}
\usepackage{algorithmic}

\usepackage{amsmath}
\usepackage{amssymb}
\usepackage{placeins}

\usepackage{multirow}

\usepackage{amsfonts}

\usepackage{newfloat}
\usepackage{listings}
\DeclareCaptionStyle{ruled}{labelfont=normalfont,labelsep=colon,strut=off} % DO NOT CHANGE THIS
\floatstyle{ruled}
\newfloat{listing}{tb}{lst}{}
\floatname{listing}{Listing}
\title{PlugTrack: Multi-Perceptive Motion Analysis for Adaptive Fusion \\ in Multi-Object Tracking}
\author{
    Seungjae Kim,
    SeungJoon Lee,
    MyeongAh Cho\thanks{Corresponding author.}
}
\affiliations{
    Department of Software Convergence, Kyung Hee University\\
    \{tmdwo8814, diplomat3334, maycho\}@khu.ac.kr
}

\begin{document}
\maketitle

% =============================================================================================
% =============================================================================================
%                                             abstract
% =============================================================================================
% =============================================================================================

\begin{abstract}
Multi-object tracking (MOT) predominantly follows the tracking-by-detection paradigm, where Kalman filters serve as the standard motion predictor due to computational efficiency but inherently fail on non-linear motion patterns. Conversely, recent data-driven motion predictors capture complex non-linear dynamics but suffer from limited domain generalization and computational overhead. Through extensive analysis, we reveal that even in datasets dominated by non-linear motion, Kalman filter outperforms data-driven predictors in up to 34\% of cases, demonstrating that real-world tracking scenarios inherently involve both linear and non-linear patterns. To leverage this complementarity, we propose PlugTrack, a novel framework that adaptively fuses Kalman filter and data-driven motion predictors through multi-perceptive motion understanding. Our approach employs multi-perceptive motion analysis to generate adaptive blending factors. PlugTrack achieves significant performance gains on MOT17/MOT20 and state-of-the-art on DanceTrack without modifying existing motion predictors. To the best of our knowledge, PlugTrack is the first framework to bridge classical and modern motion prediction paradigms through adaptive fusion in MOT. 
\end{abstract}

\begin{links}
    \link{Code}{https://github.com/VisualScienceLab-KHU/PlugTrack}
\end{links}

\begin{figure}[t]
    \centering
    \includegraphics[width=\columnwidth]{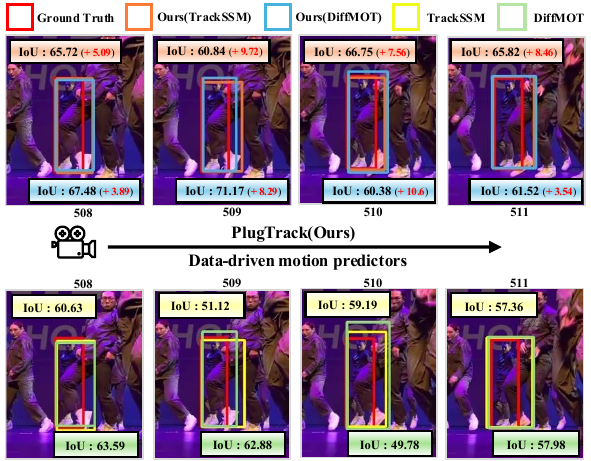} 
    \caption{Qualitative comparison of motion prediction on DanceTrack (frames 508-511) dataset. Top row: PlugTrack results using TrackSSM and DiffMOT as base predictors. Bottom row: Standalone TrackSSM and DiffMOT results. PlugTrack adaptively fuses Kalman filter and data-driven predictions to better handle both linear and non-linear motions, achieving up to +10.6 IoU gains. This demonstrates its core novelty: a lightweight, plug-in mechanism that dynamically integrates complementary motion cues to outperform individual predictors in complex scenarios.}
    \label{fig:fig1}
\end{figure}

% =============================================================================================
% =============================================================================================
%                                             Introduction
% =============================================================================================
% =============================================================================================


\section{Introduction}

Multi-object tracking (MOT) aims to detect and track multiple objects across video sequences while maintaining identity consistency. This fundamental computer vision task enables extensive applications including autonomous driving, surveillance systems, and robotics. Most MOT methods follow the tracking-by-detection paradigm, which decomposes the problem into three subtasks: object detection, motion prediction, and association between detections and predictions. Among these subtasks, motion prediction plays a critical role in maintaining object identities by accurately forecasting object movements during occlusions. Given the importance of real-time performance in MOT applications, tracking-by-detection methods have traditionally adopted the Kalman filter as their motion predictor due to its parameter-free design and computational efficiency. The Kalman filter's recursive update mechanism inherently makes linear motion assumptions, proving effective on datasets dominated by linear movements such as MOT17 ~\cite{milan2016mot16} and MOT20 ~\cite{dendorfer2020mot20}. However, this same recursive mechanism renders it ineffective on non-linear motion datasets like DanceTrack ~\cite{sun2022dancetrack} and SportsMOT ~\cite{cui2023sportsmot}, where complex movements and abrupt directional changes violate its fundamental linear assumptions.

To achieve robust tracking under non-linear motion, existing tracking-by-detection methods have pursued two primary approaches. The first approach employs Kalman filter-based methods with rule-based modifications such as OC-SORT ~\cite{cao2023observation} and Hybrid-SORT ~\cite{yang2024hybrid}. These heuristic methods ensure real-time performance but require extensive scenario-specific parameter tuning. This dependency fundamentally limits their domain generalization capabilities. The second approach replaces the Kalman filter with data-driven motion predictors such as DiffMOT ~\cite{lv2024diffmot}, which leverages diffusion models for probabilistic motion prediction, and TrackSSM ~\cite{hu2024trackssm}, which employs state-space models to capture temporal dynamics. Unlike the Kalman filter, these learnable architectures demonstrate robust performance in capturing non-linear motion. However, they tend to overfit to their training data distributions, making it challenging to adapt to diverse motion patterns across different domains. These limitations highlight the critical need for developing a motion predictor that achieves both generalization across varied motion patterns and real-time efficiency.

To understand the strengths of existing motion predictors, we conducted evaluations across both linear and non-linear motion domains. We assessed three motion predictors—Kalman filter, DiffMOT, and TrackSSM—by generating predictions on identical tracklets and measuring their IoU scores against ground truth. Figure~\ref{fig:fig2} presents the number of tracklets where each predictor achieved the highest IoU score. On the linear motion domain MOT17, the Kalman filter demonstrates clear superiority over data-driven motion predictors, achieving the best predictions on 60.3\% of all tracklets (12,061 out of 20,000). Surprisingly, even on the non-linear motion domain DanceTrack, the Kalman filter outperforms data-driven motion predictors in 34\% of all tracklets (1,700 out of 5,000). This surprising result indicates that linear motion patterns frequently occur even within datasets designed for complex non-linear movements. This observation reveals a crucial insight: \textbf{real-world tracking scenarios inherently involve both linear and non-linear motion patterns, regardless of domain-specific characteristics}. This necessitates a unified framework that can adaptively respond to both motion types rather than treating them as mutually exclusive.

Based on these observations and the limitations of existing approaches, we identify two critical requirements for a practical motion prediction system in MOT. \textbf{(1) Complementary modeling of linear and non-linear motion:} to handle the full spectrum of real-world motion dynamics, a unified approach must integrate both Kalman filter and data-driven motion predictors. \textbf{(2) Real-time capability and generalization:} the framework must be lightweight enough for real-time operation, while maintaining robustness across unseen domains and motion types.

To this end, we propose \textbf{PlugTrack}, a framework where a Kalman filter and a data-driven motion predictor work in a complementary and adaptive manner. Our approach consists of two key components: (1) the Contextual Motion Encoder (CME) that performs multi-perceptive motion analysis, and (2) the Adaptive Blending Generator (ABG) that produces adaptive blending factors based on CME's multi-perceptive understanding. These adaptive blending factors are then used to blend predictions from both the Kalman filter and data-driven motion predictor, achieving refined motion prediction. This plug-and-play design enhances existing pretrained data-driven motion predictors, while PlugTrack's lightweight architecture maintains real-time capability.

\begin{figure}[t]
    \centering
    \includegraphics[width=0.9\columnwidth]{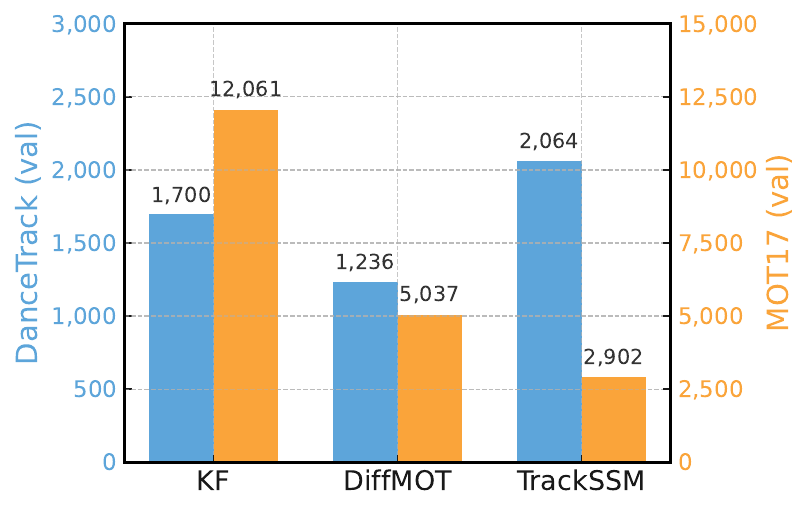} 
    \caption{Comparison of motion predictor performance on DanceTrack and MOT17, showing the number of tracklets where each predictor (Kalman filter, DiffMOT, TrackSSM) achieves the highest IoU score with ground truth.}
    \label{fig:fig2}
\end{figure}

% =============================================================================================
% =============================================================================================
%                                             Related Work
% =============================================================================================
% =============================================================================================

\begin{figure*}[t]
    \centering
    \includegraphics[width=0.9\textwidth]{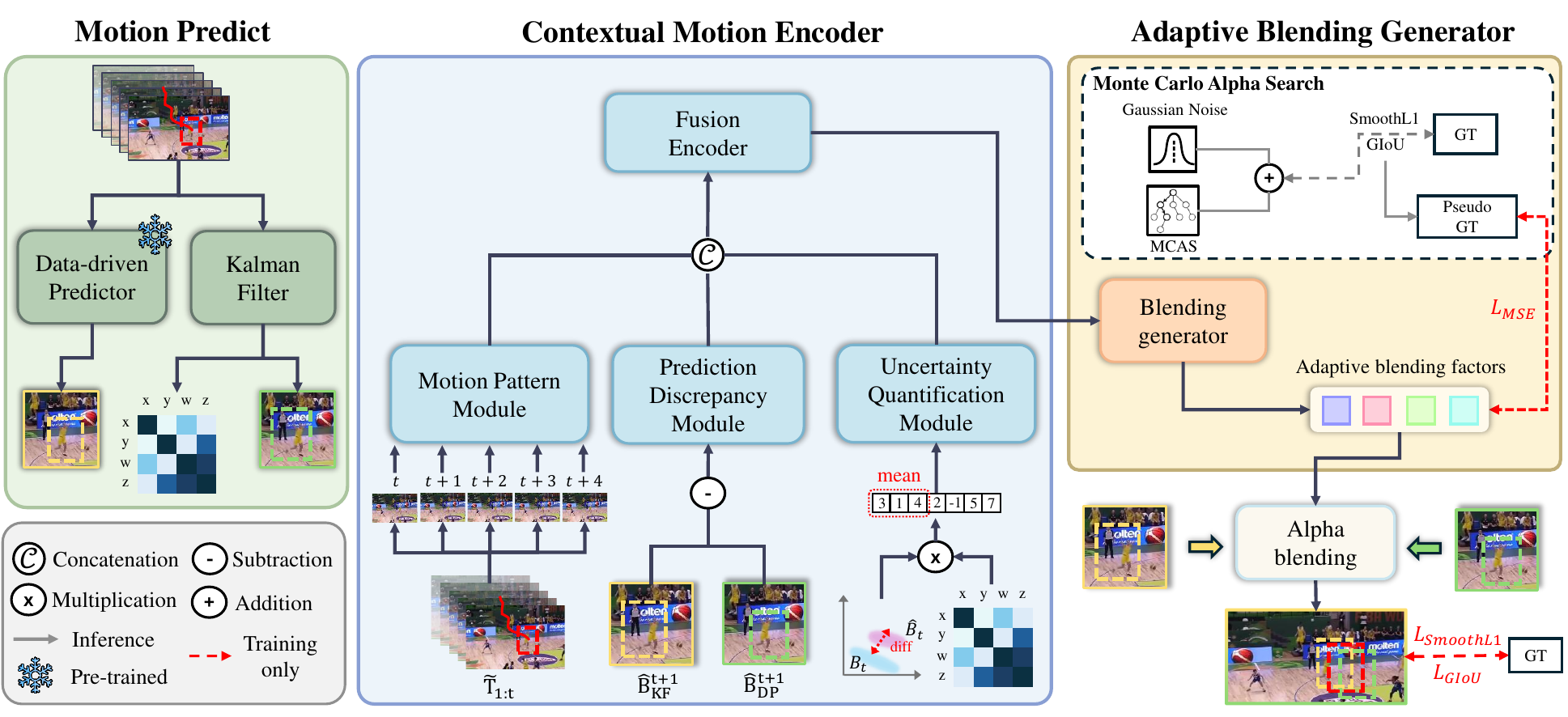} 
    \caption{Overview of the PlugTrack architecture. Our framework consists of two main components: (1) Contextual Motion Encoder (CME) that analyzes motion from multi-perceptive through three specialized modules to generate multi-perceptive motion feature. (2) Then Adaptive Blending Generator (ABG) that produces adaptive blending factors for alpha blending. During training, Monte Carlo Alpha Search (MCAS) generates pseudo ground truth blending factors by evaluating multiple candidates with added Gaussian noise. During inference, the learned ABG directly predicts optimal blending factors for real-time adaptive fusion of Kalman filter and data-driven motion predictor outputs.}
    \label{fig:fig3}
\end{figure*}

\section{Related Work}

\subsection{Kalman Filter Based Motion Prediction Methods}

The Kalman filter has dominated motion prediction in MOT due to its computational efficiency. SORT~\cite{bewley2016simple} established the tracking-by-detection paradigm, while DeepSORT~\cite{wojke2017simple} added appearance features. Recent methods attempt to address the Kalman filter's inability to handle non-linear motion through heuristic modifications: OC-SORT~\cite{cao2023observation} introduces observation-centric recovery, C-BIoU~\cite{yang2023hard} proposes buffered matching, and Hybrid-SORT~\cite{yang2024hybrid} combines multiple strategies. However, these approaches share critical limitations: (1) they require extensive manual tuning for different scenarios, (2) their rule-based modifications cannot overcome the Kalman filter's fundamental linear motion assumption, and (3) they lack adaptability to varying motion complexities within sequences. While computationally efficient, these methods have limited domain generalization capabilities and cannot adapt to diverse motion patterns, particularly failing on datasets with complex movements like DanceTrack.

\subsection{Data-driven Motion Prediction Methods}

Data-driven approaches address the limitations of the Kalman filter's linear motion assumption by capturing complex motion patterns. TrackSSM~\cite{hu2024trackssm} and DiffMOT~\cite{lv2024diffmot} replace the Kalman filter with state-space models and diffusion-based models respectively for motion prediction. Despite their superior performance on non-linear benchmarks, data-driven predictors introduce several challenges: (1) Domain overfitting: Models trained on complex motions struggle with simple linear trajectories, (2) Computational overhead: Methods like MOTIP~\cite{gao2025multiple} sacrifice real-time capability for accuracy, and (3) Black-box nature: Unlike the interpretable nature of Kalman filters, neural predictors offer limited transparency regarding their failure modes.

The critical gap is the false dichotomy between Kalman filter and data-driven motion predictors. Current approaches treat them as mutually exclusive, missing their complementary strengths. Real-world tracking scenarios inherently contain both linear and non-linear motion patterns that no single approach can optimally handle. While adaptive methods like RelationTrack~\cite{yu2022relationtrack} and StrongSORT~\cite{du2023strongsort} explore fusion strategies, they operate within homogeneous paradigms. None bridge the divide between Kalman filter and data-driven approaches, necessitating an adaptive framework that intelligently blends both based on motion context.

% =============================================================================================
% =============================================================================================
%                                             Method
% =============================================================================================
% =============================================================================================

\section{Proposed Method}

We present PlugTrack, a novel framework that adaptively fuses the Kalman filter and data-driven motion predictors. As shown in Figure~\ref{fig:fig3}, our approach consists of two core components. The Contextual Motion Encoder (CME) extracts multi-perceptive motion features $\mathbf{f}_{mult}$ by analyzing tracklet patterns, prediction discrepancies, and uncertainty measures. This multi-perceptive analysis enables CME to capture nuanced motion characteristics that single predictors might miss. The Adaptive Blending Generator (ABG) then converts $\mathbf{f}_{mult}$ into adaptive blending factors $\alpha$, which determine the optimal contribution of each predictor based on the current motion context. These factors dynamically weight the predictions from both motion predictors to produce the final bounding box. PlugTrack's plug-and-play design allows seamless integration with any existing data-driven motion predictor without architectural modifications.

\subsection{Contextual Motion Encoder}

To formally characterize the input space of our method, we define the following representations. Let $\tilde{\mathbf{T}}_{1:t} = \{\mathbf{B}_1, \ldots, \mathbf{B}_f, \ldots, \mathbf{B}_t\}$ denote the tracklet up to time~$t$, where each $\mathbf{B}_f$ represents the detection result at frame $f$. Each bounding box $\mathbf{B}_f = (x_f, y_f, w_f, h_f, \Delta x_f, \Delta y_f, \Delta w_f, \Delta h_f)$ contains both position and velocity information, where $\Delta$ denotes the difference from frame $f-1$ to $f$. The Kalman filter and data-driven motion predictor generate predictions $\hat{\mathbf{B}}_{t+1}^{KF} \in \mathbb{R}^4$ and $\hat{\mathbf{B}}_{t+1}^{DP} \in \mathbb{R}^4$, respectively. Additionally, we extract motion uncertainty $\boldsymbol{\sigma}_{KF} \in \mathbb{R}^4$ from the Kalman filter's internal state estimates.

The CME consists of three specialized modules that examine different aspects of motion patterns:

\noindent{\textbf{(1) Motion Pattern Module.}}
As the first component of multi-perceptive motion analysis, the Motion Pattern Module (MPM) captures temporal dependencies and motion characteristics from the tracklet. We employ a Long Short-Term Memory network~\cite{graves2012long} to encode the sequential motion information:
\begin{equation}
\mathbf{f}_{MPM} = \operatorname{LSTM}(\tilde{\mathbf{T}}_{1:t}) = \mathbf{h}_t
\end{equation}
where $\mathbf{h}_t \in \mathbb{R}^{128}$ represents the final hidden state of the LSTM. This design enables MPM to learn complex motion patterns such as acceleration, deceleration, and directional changes.

\noindent{\textbf{(2) Prediction Discrepancy Module.}}
As the second component of multi-perceptive motion analysis, the Prediction Discrepancy Module (PDM) quantifies and interprets the differences between the Kalman filter and data-driven motion predictor predictions. We compute the prediction discrepancy as $\mathbf{B}_{disc} = \hat{\mathbf{B}}_{t+1}^{KF} - \hat{\mathbf{B}}_{t+1}^{DP}$, which is then processed through multiple linear layers:
\begin{equation}
\mathbf{f}_{PDM} = \operatorname{MLP}(\mathbf{B}_{disc})
\end{equation}
where $\mathbf{f}_{PDM} \in \mathbb{R}^{32}$. The PDM learns to interpret this discrepancy as indicators of motion complexity and predictor reliability. A large discrepancy often signals transitions between linear and non-linear motion phases. In the subsequent fusion process, $\mathbf{f}_{PDM}$ is fused with $\mathbf{f}_{MPM}$. This fusion enables the model to contextualize prediction differences within specific motion scenarios, providing valuable cues for determining when to trust each predictor.

\noindent{\textbf{(3) Uncertainty Quantification Module.}}
As the third component of multi-perceptive motion analysis, the Uncertainty Quantification Module (UQM) leverages the uncertainty measure extracted from the Kalman filter's internal state estimation process. We compute dimension-wise Normalized Innovation Squared (NIS) to quantify prediction confidence for each bounding box dimension:
\begin{equation}
\operatorname{NIS}_{t,i} = \frac{(B_{t,i} - \hat{B}_{t,i})^2}{S_{t,ii}}
\end{equation}
where $B_{t,i} - \hat{B}_{t,i}$ is the innovation (prediction error) for dimension $i \in \{x, y, w, h\}$, and $S_{t,ii}$ is the corresponding diagonal element of the innovation covariance matrix $\mathbf{S}_t$. Please refer to Appendix for more details.

To obtain a robust uncertainty measure, we aggregate recent NIS values by summing their mean and standard deviation, capturing not only the average prediction error but also its consistency over time. This yields a 4-dimensional uncertainty vector $\boldsymbol{\sigma}_{KF} \in \mathbb{R}^4$ that comprehensively characterizes the Kalman filter's confidence across all bounding box dimensions.

This uncertainty vector is processed through an MLP to extract motion-relevant features:
\begin{equation}
\mathbf{f}_{UQM} = \operatorname{MLP}(\boldsymbol{\sigma}_{KF})
\end{equation}
where $\mathbf{f}_{UQM} \in \mathbb{R}^{32}$. The UQM provides an objective measure of the Kalman filter's prediction reliability. High uncertainty values indicate that the Kalman filter has low confidence in its predictions, suggesting the presence of non-linear motion patterns that violate its linear assumptions. This uncertainty-based motion characterization enables the model to implicitly distinguish between linear and non-linear motion phases for adaptive blending.

The outputs from the three modules, each providing a distinct component of multi-perceptive analysis, are subsequently fused to create a multi-perceptive motion feature $\mathbf{f}_{mult}$:
\begin{equation}
\mathbf{f}_{mult} = \operatorname{Encoder}(\operatorname{Concat}(\mathbf{f}_{MPM},\mathbf{f}_{PDM},\mathbf{f}_{UQM}))
\end{equation}
This multi-perceptive analysis enables CME to capture the nuanced relationships between motion patterns, prediction differences, and uncertainty levels, providing rich information for subsequent adaptive blending decisions.

\subsection{Adaptive Blending Generator}

The ABG transforms the multi-perceptive motion feature $\mathbf{f}_{mult}$ into adaptive blending factors $\tilde{\boldsymbol{\alpha}} = \operatorname{MLP}(\mathbf{f}_{mult})$. The resulting $\tilde{\boldsymbol{\alpha}}$ is a 4-dimensional vector containing blending factors for each bounding box coordinate (x, y, w, h). The ABG is implemented as an MLP with sigmoid activation in the final layer, ensuring all blending factors are bounded in range $[0,1]$.

The final bounding box prediction synthesizes both motion predictions through element-wise weighted blending:
\begin{equation}
\hat{\mathbf{B}}_{ABG} = \tilde{\boldsymbol{\alpha}} \odot \hat{\mathbf{B}}_{KF} + (1-\tilde{\boldsymbol{\alpha}}) \odot \hat{\mathbf{B}}_{DP}
\end{equation}
where $\tilde{\boldsymbol{\alpha}} = (\alpha_x, \alpha_y, \alpha_w, \alpha_h)$ contains the adaptive blending factors. This coordinate-specific formulation allows ABG to dynamically adjust the predictions of each predictor based on the motion characteristics identified by CME. For instance, when horizontally linear motion occurs, CME detects stable horizontal  patterns through MPM while UQM indicates low uncertainty, then ABG will assign higher weight to the Kalman filter for x-coordinate ($\alpha_x > 0.5$). Conversely, when vertically non-linear motion occurs, PDM reveals large prediction discrepancies during rapid vertical movements, ABG will rely more on the data-driven motion predictor for y-coordinate ($\alpha_y < 0.5$). This adaptive strategy, guided by CME's multi-perceptive analysis, effectively handles complex motion scenarios without manual intervention.

% =============================================================================================
                                    % Table 1 : mot17, mot20 
\begin{table*}[t!]
\centering
\begin{tabular}{l|ccccc|ccccc}
\hline
\multicolumn{1}{c}{Method}
  & \multicolumn{5}{c|}{MOT17}
  & \multicolumn{5}{c}{MOT20} \\
\hline
  & HOTA & IDF1 & AssA & DetA & MOTA
  & HOTA & IDF1 & AssA & DetA & MOTA \\
\hline
\multicolumn{11}{l}{\textit{\textbf{Kalman Filter Based}}} \\
FairMOT      
             & 59.3 & 72.3 & 58.0 & 60.9 & 73.7
             & 54.6 & 67.3 & 54.7 & 54.7 & 61.8 \\
ByteTrack    
             & 63.1 & 77.3 & 62.0 & 64.5 & \textbf{80.3}
             & 61.3 & 75.2 & 59.6 & \textbf{63.4} & \textbf{77.8} \\
OC-SORT      
             & 63.2 & 77.5 & 63.4 & 63.2 & 78.0
             & 62.4 & \textbf{76.3} & \textbf{62.5} & 62.4 & 75.7 \\
C-BIoU       
             & 64.1 & \textbf{79.7} & 63.7 & \textbf{64.8} & 79.7
             & / & / & / & / & / \\
Hybrid-SORT  
             & 63.6 & 78.4 & / & / & 79.3
             & \textbf{62.5} & 76.2 & / & / & 76.4 \\
\hline
\multicolumn{11}{l}{\textit{\textbf{Data-driven Based}}} \\
DiffusionTrack   
                 & 60.8 & 73.8 & 58.8 & 63.2 & 77.9
                 & 55.3 & 66.3 & 51.3 & 59.9 & 72.8 \\
ETTrack          
                 & 61.9 & 75.9 & 60.5 & / & 79.0
                 & / & / & / & / & / \\
\textbf{TrackSSM}
                 & 61.4 & 74.1 & 59.6 & 63.6 & 78.5
                 & 59.1 & 71.1 & 57.5 & 60.9 & 73.9 \\
\textbf{DiffMOT}
                 & 64.0 & 78.9 & 64.2 & 64.1 & 79.1
                 & 61.6 & 74.9 & 60.5 & 62.8 & 76.3 \\
\hline
\textbf{Ours(TrackSSM)}
  & \shortstack[c]{61.9 \\ (+0.5)}
  & \shortstack[c]{75.2 \\ (+1.1)}
  & \shortstack[c]{60.3 \\ (+0.7)}
  & \shortstack[c]{63.9 \\ (+0.3)}
  & \shortstack[c]{78.7 \\ (+0.1)}
  & \shortstack[c]{59.7 \\ (+0.6)}
  & \shortstack[c]{72.3 \\ (+1.2)}
  & \shortstack[c]{58.5 \\ (+1.0)}
  & \shortstack[c]{61.3 \\ (+0.4)}
  & \shortstack[c]{74.5 \\ (+0.6)} \\
\textbf{Ours(DiffMOT)}
  & \shortstack[c]{\textbf{64.2} \\ (+0.2)}
  & \shortstack[c]{79.0 \\ (+0.1)}
  & \shortstack[c]{\textbf{64.4} \\ (+0.2)}
  & \shortstack[c]{64.0 \\ (-0.1)}
  & \shortstack[c]{79.2 \\ (+0.1)} 
  & \shortstack[c]{61.8 \\ (+0.2)}
  & \shortstack[c]{75.2 \\ (+0.3)}
  & \shortstack[c]{60.9 \\ (+0.4)}
  & \shortstack[c]{62.9 \\ (+0.1)}
  & \shortstack[c]{76.4 \\ (+0.1)}\\
\hline
\end{tabular}

\caption{Results on MOT17 \& MOT20 test. The best results are shown in bold.}
\label{tab:table1}
\end{table*}

% =============================================================================================

\subsection{Monte Carlo Alpha Search}

Training adaptive blending factors through direct optimization presents a fundamental challenge: the ABG tends to converge to dataset-specific biases rather than learning to adaptively respond to varying motion dynamics. Specifically, when trained on MOT17 where linear motion dominates, the ABG consistently assigns high weights to the Kalman filter's predictions across all scenarios, failing to recognize cases where the data-driven motion predictor would be more appropriate. This \textit{bias collapse} problem prevents the model from discovering the complementary strengths of both predictors.

To address this challenge, we propose Monte Carlo Alpha Search (MCAS), inspired by the success of Monte Carlo methods in navigating complex optimization landscapes. Similar to how Monte Carlo methods in ~\cite{hampali2021monte}  explore discrete proposals in 3D scene understanding by systematically evaluating candidate solutions, MCAS searches through a discrete space of blending factor combinations to find optimal fusion strategies for each training sample.

We define a discrete search space of blending factors:
\begin{equation}
\mathcal{A} = \{0.3, 0.4, 0.5, 0.6, 0.7\}^4
\end{equation}
This creates $5^4 = 625$ candidate combinations, constraining the search within $[\lambda_1, \lambda_2] = [0.3, 0.7]$ to ensure meaningful contributions from both predictors.

For each training batch, we introduce stochastic exploration by adding Gaussian noise:
\begin{equation}
\tilde{\mathcal{A}}_b = \operatorname{clamp}(\mathcal{A} + \epsilon_b, 0, 1), \quad \epsilon_b \sim \mathcal{N}(0, 0.1^2)
\end{equation}
where the noise is applied independently to each candidate and batch, encouraging diverse exploration patterns.

Given predictions $\hat{\mathbf{B}}^{KF}$ and $\hat{\mathbf{B}}^{DP}$, we evaluate each candidate $\boldsymbol{\alpha} \in \tilde{\mathcal{A}}_b$ by computing the blended prediction and measuring its accuracy:
\begin{equation}
\mathcal{S}(\boldsymbol{\alpha}) = \operatorname{SmoothL1}(\hat{\mathbf{B}}(\boldsymbol{\alpha}), \mathbf{B}_{GT}) + \operatorname{GIoU}(\hat{\mathbf{B}}(\boldsymbol{\alpha}), \mathbf{B}_{GT})
\end{equation}
where $\hat{\mathbf{B}}(\boldsymbol{\alpha}) = \boldsymbol{\alpha} \odot \hat{\mathbf{B}}^{KF} + (1-\boldsymbol{\alpha}) \odot \hat{\mathbf{B}}^{DP}$.

The optimal blending factors for each sample are selected as:
\begin{equation}
\boldsymbol{\alpha}^* = \arg\min_{\boldsymbol{\alpha} \in \tilde{\mathcal{A}}_b} \mathcal{S}(\boldsymbol{\alpha})
\end{equation}

These pseudo ground-truth values $\boldsymbol{\alpha}^*$ serve as supervision signals for training the ABG through an auxiliary loss:
\begin{equation}
\mathcal{L}_{\text{MCAS}} = \operatorname{MSE}(\tilde{\boldsymbol{\alpha}}, \boldsymbol{\alpha}^*)
\end{equation}

By providing explicit supervision for optimal blending strategies, MCAS guides the ABG to learn context-appropriate fusion rather than dataset-specific biases. During inference, MCAS is not required as the trained ABG directly predicts adaptive blending factors, maintaining real-time efficiency.

\subsection{Training loss}

The overall training objective of PlugTrack is formulated as:
\begin{equation}
\mathcal{L} = \mathcal{L}_{\text{SmoothL1}} + \mathcal{L}_{\text{GIoU}} + \mathcal{L}_{\text{MCAS}}
\end{equation}
where $\mathcal{L}_{\text{SmoothL1}}$ and $\mathcal{L}_\text{{GIoU}}$ measure the accuracy of the final blended prediction $\hat{\mathbf{B}}_{ABG}$ against ground truth, while $\mathcal{L}_{\text{MCAS}}$ provides auxiliary supervision for learning optimal blending strategies.

% =============================================================================================
% =============================================================================================
%                                             Experiments
% =============================================================================================
% =============================================================================================

\section{Experiments}

\subsection{Experimental Setup}

\noindent\textbf{Datasets.}
We evaluate our method on three widely-used MOT benchmarks: MOT17~\cite{milan2016mot16}, MOT20~\cite{dendorfer2020mot20}, and DanceTrack~\cite{sun2022dancetrack}. MOT17 and MOT20 predominantly feature pedestrians with linear motion patterns in surveillance scenarios. DanceTrack contains dancers performing complex non-linear movements with frequent occlusions and similar appearances. For training on linear motion domains, we combine MOT17 and MOT20 training sets to create a mixed dataset, which provides diverse linear motion patterns along with crowded scenes.

\noindent\textbf{Evaluation Metrics.}
We adopt the standard MOT evaluation metrics for comprehensive assessment. HOTA~\cite{luiten2021hota} provides a balanced measure of detection and association performance. IDF1 emphasizes identity preservation across frames. AssA and DetA decompose HOTA into association and detection components respectively. MOTA measures overall tracking accuracy considering false positives, false negatives, and identity switches.

\noindent\textbf{Implementation Details.}
Following the tracking-by-detection paradigm, we employ YOLOX~\cite{ge2021yolox} as our detector for all experiments except FairMOT~\cite{zhang2021fairmot} comparisons, ensuring fair comparison across methods. Our framework is trained using Adam optimizer with a learning rate of 0.001. We train for 270 epochs on MIX(MOT17\&20) dataset and 220 epochs on DanceTrack without any data augmentation or sampling techniques. The batch size is set to 2,048. We use fixed-length tracklets of 5 frames as the model's input. The LSTM encoder consists of 2 layers with hidden dimension 128. For the Uncertainty Quantification Module, we use a sliding window size of $w$=3 for computing the averaged NIS values. For MCAS, we set the alpha range bounds $\lambda_1$=0.3 and $\lambda_2$=0.7, with Gaussian noise standard deviation of 0.1 for stochastic exploration. Both TrackSSM and DiffMOT are initialized with their official pre-trained weights. All experiments are conducted on a single NVIDIA GeForce RTX 4090 GPU.

%=============================================================================================
                                    % Table 2 : DanceTrack

\begin{table}[htbp]
\centering
\small
\setlength{\tabcolsep}{4.5pt} 
\begin{tabular}{lccccc}
\hline
\multicolumn{1}{c}{Method}
  & \multicolumn{5}{c}{DanceTrack} \\
\hline
  & HOTA & IDF1 & AssA & DetA & MOTA \\
\hline
\multicolumn{6}{l}{\textit{\textbf{Kalman Filter Based}}} \\
FairMOT      & 39.7 & 40.8 & 23.8 & 66.7 & 82.2 \\
ByteTrack    & 47.3 & 52.5 & 31.4 & 71.6 & 89.5 \\
OC-SORT      & 55.1 & 54.2 & 38.0 & 80.3 & 89.4 \\
C-BIoU       & 60.6 & 61.6 & 45.4 & 81.3 & 91.6 \\
Hybrid-SORT  & 62.2 & 63.0 & / & / & 91.6 \\
\hline
\multicolumn{6}{l}{\textit{\textbf{Data-driven Based}}} \\
DiffusionTrack   & 52.4 & 47.5 & 33.5 & 82.2 & 89.3 \\
ETTrack          & 56.4 & 57.5 & 39.1 & 81.7 & 92.2 \\
MambaTrack       & 56.8 & 57.8 & 39.8 & 80.1 & 90.1 \\
\textbf{TrackSSM}& 57.7 & 57.5 & 41.0 & 81.5 & 92.2 \\
MotionTrack      & 58.2 & 58.6 & 41.7 & 81.4 & 91.3 \\
\textbf{DiffMOT} & 62.3 & 63.0 & 47.2 & \textbf{82.5} & \textbf{92.8} \\
\hline
\textbf{Ours(TrackSSM)}
  & \shortstack[c]{59.2 \\ (+1.5)}
  & \shortstack[c]{59.0 \\ (+1.5)}
  & \shortstack[c]{42.9 \\ (+1.9)}
  & \shortstack[c]{81.9 \\ (+0.4)}
  & \shortstack[c]{92.2 \\ (0)} \\
\textbf{Ours(DiffMOT)}
  & \shortstack[c]{\textbf{63.3} \\ (+1.0)}
  & \shortstack[c]{\textbf{64.1} \\ (+1.1)}
  & \shortstack[c]{\textbf{48.4} \\ (+1.2)}
  & \shortstack[c]{\textbf{82.5} \\ (0)}
  & \shortstack[c]{92.4 \\ (-0.4)} \\
\hline
\end{tabular}

\caption{Results on DanceTrack test set.  The best results in bold.}
\label{tab:table2}
\end{table}

%=============================================================================================

\subsection{Benchmark Evaluation}

\noindent\textbf{Performance on Linear Motion Datasets.}
To evaluate PlugTrack's ability to enhance data-driven predictors on linear motion, we test on MOT17 and MOT20 datasets. Table~\ref{tab:table1} shows Kalman filter-based methods achieve state-of-the-art performance across most metrics, consistent with Figure~\ref{fig:fig2} where Kalman filter dominates data-driven predictors in linear scenarios. This explains the strong performance of C-BIoU and Hybrid-SORT, which augment Kalman filter with heuristics. Despite this, PlugTrack improves base predictors: Ours(TrackSSM) gains 0.5/0.6 HOTA on MOT17/MOT20, while Ours(DiffMOT) improves by 0.2/0.2. Notably, Ours(DiffMOT) achieves state-of-the-art HOTA (64.2) and AssA (64.4) on MOT17, demonstrating our adaptive fusion can surpass pure Kalman methods. These results validate that our framework leverages Kalman filter's linear motion strength while maintaining data-driven flexibility.

\noindent\textbf{Performance on Non-linear Motion Dataset.}
We further assess whether adaptive fusion improves tracking on complex non-linear motion. Table~\ref{tab:table2} presents DanceTrack results, where PlugTrack demonstrates substantial improvements: HOTA gains of 1.5 and IDF1 of 1.5 for TrackSSM, and 1.0 and 1.1 for DiffMOT. The significant AssA gains indicate effective identity maintenance during challenging motion transitions. PlugTrack(DiffMOT) achieves state-of-the-art performance with 63.3 HOTA and 48.4 AssA, surpassing specialized methods like MotionTrack and MambaTrack. These results confirm that intelligently combining Kalman filter and data-driven predictors yields superior performance on non-linear motion-dominant datasets.

\subsection{Cross-domain Generalization}

To investigate PlugTrack's generalization across fundamentally different motion domains, we conduct bidirectional transfer experiments. Table~\ref{tab:table3} shows cross-domain results using DanceTrack validation and MOT20 test sets. When trained on DanceTrack and tested on MOT20, PlugTrack(TrackSSM) achieves remarkable improvements: HOTA +6.5, IDF1 +9.2, and AssA +8.7. This large gain occurs because MOT20 contains many linear motion patterns where Kalman filter naturally excels, as shown in Figure~\ref{fig:fig2}. Our CME correctly identifies these linear patterns, and ABG assigns higher weights to the Kalman filter's predictions. Conversely, when trained on MOT20 and tested on DanceTrack, PlugTrack still improves performance: HOTA +1.8, IDF1 +3.3, and AssA +2.3. These gains are smaller but significant—both the Kalman filter and linear-domain-trained predictor face extreme challenges in non-linear domains. That our framework still finds optimal blending strategies in this adverse setting demonstrates the strength of multi-perceptive motion analysis, proving PlugTrack can extract meaningful improvements even when both predictors are mismatched to the target domain.

% =============================================================================================
                                % Table 3 : Cross-Domain

\begin{table}[]
\centering
\small
\begin{tabular}{lccc}
\hline
Model & HOTA & IDF1 & AssA \\ 
\hline
\multicolumn{4}{l}{\textbf{\textit{DanceTrack $\rightarrow$ MOT20}}} \\
\hline
TrackSSM & 47.7 & 56.0 & 41.7 \\ 
PlugTrack(TrackSSM) & \textbf{54.2} & \textbf{65.2} & \textbf{50.4} \\ 
\hline
\multicolumn{4}{l}{\textbf{\textit{MOT20 $\rightarrow$ DanceTrack}}} \\
\hline
TrackSSM & 52.5 & 49.0 & 34.8 \\ 
PlugTrack(TrackSSM) & \textbf{54.3} & \textbf{52.3} & \textbf{37.1} \\ 
\hline
\end{tabular}
\caption{Cross-domain Performance Comparison}
\label{tab:table3}
\end{table}

% =============================================================================================

\subsection{Efficiency Analysis}

Real-time performance is critical for practical MOT applications. Table 4 demonstrates that PlugTrack maintains real-time capability while achieving substantial tracking improvements. Our lightweight design adds only 0.54M parameters to both base predictors, representing a 22\% increase for TrackSSM and 4.7\% for DiffMOT. The resulting processing speeds remain practical: PlugTrack(TrackSSM) operates at 34.2 FPS (down from 37.2 FPS) and PlugTrack(DiffMOT) at 24.7 FPS (down from 25.9 FPS), both exceeding the 20 FPS real-time threshold. This efficiency stems from our compact architecture—CME uses a 128-dimensional LSTM with lightweight MLPs, while ABG consists of simple fully-connected layers. These results validate that our multi-perceptive motion analysis introduces minimal computational overhead, successfully balancing enhanced tracking accuracy with real-time constraints.

% =============================================================================================
                                % Table 4 : efficiency of ours model

\begin{table}[h!]
\centering
\small
\begin{tabular}{l|c|c}
\hline
Model & Params (M) & FPS \\
\hline
TrackSSM & 2.5 & 37.2 \\
PlugTrack(TrackSSM) & 3.04 & 34.2 \\
\hline
DiffMOT & 11.52 & 25.9 \\
PlugTrack(DiffMOT) & 12.06 & 24.7 \\
\hline
\end{tabular}
\caption{Efficiency of PlugTrack}
\label{tab:table4}
\end{table}

% =============================================================================================

\subsection{Ablation Study}

\noindent\textbf{Component Analysis.}
To understand each component's contribution to multi-perceptive analysis, we conduct systematic ablations. Table~\ref{tab:table5} analyzes the contribution of each CME module on DanceTrack validation set. The baseline configuration (first row) directly feeds the tracklet $\tilde{\mathbf{T}}_{1:t}$ to ABG without specialized modules, achieving 59.2 HOTA. Adding only Motion Pattern Module (MPM) improves HOTA by 1.0 to 60.2, demonstrating temporal pattern analysis value. Combining MPM with PDM increases performance to 60.4 HOTA, while MPM with UQM achieves 60.3 HOTA. The full configuration with all three modules achieves the best performance of 60.8 HOTA, with AssA improving from 44.5 to 46.6. This demonstrates that multi-perceptive motion analysis creates synergistic interactions for enhanced tracking accuracy.

\noindent\textbf{Alpha Range Analysis.}
Table~\ref{tab:table6} investigates the impact of alpha range bounds in MCAS. We evaluate four range settings: [0.1, 0.9], [0.2, 0.8], [0.3, 0.7], and [0.4, 0.6]. For PlugTrack(TrackSSM), the optimal range [0.3, 0.7] achieves 54.9 HOTA, while PlugTrack(DiffMOT) performs best with [0.3, 0.7] reaching 61.7 HOTA. Wider ranges like [0.1, 0.9] allow extreme blending values that may completely ignore one predictor, leading to suboptimal performance. Conversely, narrow ranges like [0.4, 0.6] restrict the adaptive capability. The consistent optimal range across different base predictors suggests that [0.3, 0.7] provides sufficient flexibility for adaptive blending while ensuring meaningful contributions from both predictors.

\subsection{Qualitative Analysis}

\begin{figure}[t]
    \centering
    \includegraphics[width=\columnwidth]{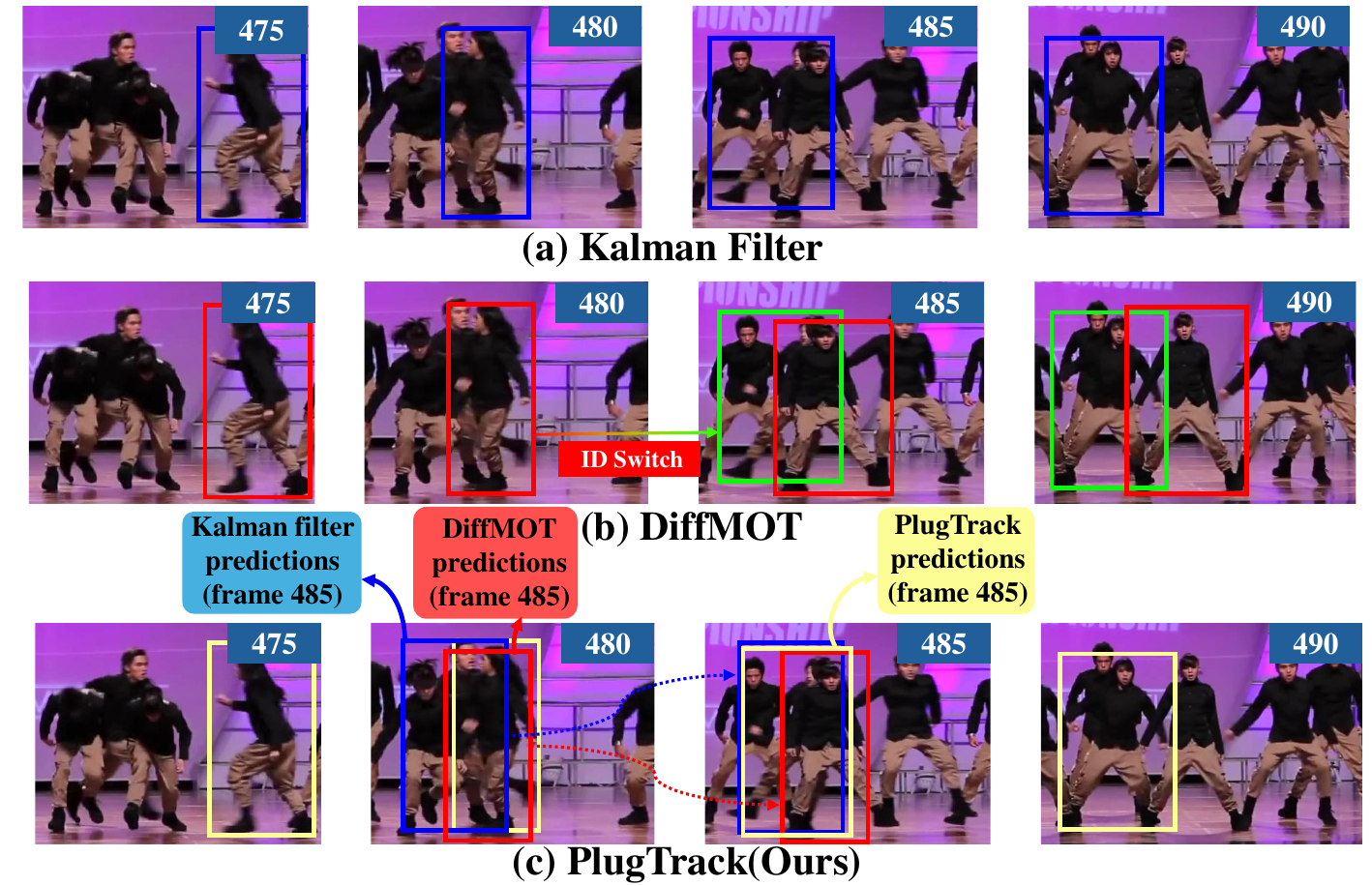} 
    \caption{Qualitative comparison on DanceTrack dataset showing tracking results across frames 475-490. (a) Kalman filter maintains ID consistency. (b) DiffMOT suffers from ID switching at frame 485. (c) PlugTrack (Ours) successfully maintains ID consistency through adaptive blending($\alpha_x$=0.874, $\alpha_y$=0.413) Boxes of the same color indicate the same tracked identity.}
    \label{fig:fig4}
\end{figure}

Finally, we perform a case analysis by comparing PlugTrack with individual motion predictors on a challenging DanceTrack sequence. Figure~\ref{fig:fig4} demonstrates PlugTrack's effectiveness in maintaining identity consistency through adaptive fusion. The visualization compares tracking results for the same object across frames 475-490, where the Kalman filter (top) successfully maintains ID, DiffMOT (middle) suffers from ID switching at frame 485, and PlugTrack (bottom) prevents this failure. Even on this non-linear motion dataset, certain scenarios favor the Kalman filter's predictions over data-driven methods. At frame 485, our adaptive blending factors ($\alpha_x=0.874$, $\alpha_y=0.413$, $\alpha_w=0.721$, $\alpha_h=0.912$) show PlugTrack assigns higher weights to the Kalman filter for most coordinates, particularly x-dimension and box size, with only y-coordinate receiving lower weight (0.413) due to non-linear vertical movement. This coordinate-specific adaptation demonstrates that CME correctly identifies motion contexts where the Kalman filter excels, enabling ABG to prevent ID switches through intelligent fusion.

% =============================================================================================
                                % Table 5 : ablation of analyzers
                                
\begin{table}[t]
\centering
\small
\begin{tabular}{c|c|c|ccc}
\hline
MPM & PDM & UQM & HOTA & AssA  & IDF1  \\
\hline
\multicolumn{6}{c}{\textbf{Ours (DiffMOT)}} \\
\hline
           &            &            & 59.2 & 44.5 & 59.7 \\
\checkmark &            &            & 60.2 & 45.8 & 61.2 \\
\checkmark & \checkmark &            & 60.4 & 46.0 & \textbf{61.8} \\
\checkmark &            & \checkmark & 60.3 & 46.1 & 61.4 \\
\checkmark & \checkmark & \checkmark & \textbf{60.8} & \textbf{46.6} & 61.7 \\
\hline
\end{tabular}
\caption{Ablation study of CME Analyzers on DanceTrack validation set. The best results are shown in bold.}
\label{tab:table5}
\end{table}

% =============================================================================================

% =============================================================================================
                                % Table 6 : ablation of MCAS ranges
                                
\begin{table}[t]
\centering
\small
\setlength{\tabcolsep}{5pt}
\begin{tabular}{l|l|ccc}
\hline
Model & Range & HOTA & IDF1 & AssA \\
\hline
\multirow{4}{*}{Ours (TrackSSM)} 
& (0.1 ~ 0.9) & 54.1 & 37.2 & 53.8 \\
& (0.2 ~ 0.8) & 54.7 & 38.1 & \textbf{54.8} \\
& (0.3 ~ 0.7) & \textbf{54.9} & \textbf{38.4} & 54.7 \\
& (0.4 ~ 0.6) & 54.8 & 38.2 & 54.7 \\
\hline
\multirow{4}{*}{Ours (DiffMOT)} 
& (0.1 ~ 0.9) & 59.8 & 45.2 & 60.7 \\
& (0.2 ~ 0.8) & 60.0 & 45.5 & 60.9 \\
& (0.3 ~ 0.7) & \textbf{60.7} & \textbf{46.5} & \textbf{61.7} \\
& (0.4 ~ 0.6) & 59.5 & 44.6 & 60.5 \\
\hline
\end{tabular}
\caption{Ablation study of different $\alpha$ ranges on DanceTrack validation set. The best results are shown in bold.}
\label{tab:table6}
\end{table}

% =============================================================================================

% =============================================================================================
                                % figure4 : qualitative
                                
% =============================================================================================
% =============================================================================================

\FloatBarrier

% =============================================================================================
% =============================================================================================
%                                             Conclusion
% =============================================================================================
% =============================================================================================

\section{Conclusion}

In this paper, we presented PlugTrack, a novel framework that adaptively fuses Kalman filter and data-driven motion predictors for robust multi-object tracking. Our key insight—that Kalman filters outperform data-driven predictors in up to 40\% of cases even on non-linear motion datasets—reveals the inherent coexistence of diverse motion patterns in real-world scenarios. Through the Contextual Motion Encoder's multi-perceptive analysis and the Adaptive Blending Generator's coordinate-specific fusion, PlugTrack achieves consistent performance improvements across MOT17, MOT20, and DanceTrack while maintaining real-time capability. The framework's plug-and-play design opens promising future directions: beyond integrating data-driven predictors, PlugTrack can potentially accommodate Kalman filter-based heuristic methods like OC-SORT and Hybrid-SORT, enabling adaptive fusion among multiple motion prediction paradigms. We believe this principle of intelligently combining classical and modern approaches provides a valuable paradigm for advancing multi-object tracking and other vision tasks where complementary methods exist.

% =============================================================================================
%                                             Acknowledgements
% =============================================================================================
\section*{Acknowledgments}
This work was supported by the National Research Foundation of Korea (NRF) grant funded by the Korea government(MSIT)(RS-2024-00456589)

\bibliography{aaai2026}

@article{milan2016mot16,
  title={MOT16: A benchmark for multi-object tracking},
  author={Milan, Anton and Leal-Taix{\'e}, Laura and Reid, Ian and Roth, Stefan and Schindler, Konrad},
  journal={arXiv preprint arXiv:1603.00831},
  year={2016}
}

@article{dendorfer2020mot20,
  title={Mot20: A benchmark for multi object tracking in crowded scenes},
  author={Dendorfer, Patrick and Rezatofighi, Hamid and Milan, Anton and Shi, Javen and Cremers, Daniel and Reid, Ian and Roth, Stefan and Schindler, Konrad and Leal-Taix{\'e}, Laura},
  journal={arXiv preprint arXiv:2003.09003},
  year={2020}
}

@inproceedings{sun2022dancetrack,
  title={Dancetrack: Multi-object tracking in uniform appearance and diverse motion},
  author={Sun, Peize and Cao, Jinkun and Jiang, Yi and Yuan, Zehuan and Bai, Song and Kitani, Kris and Luo, Ping},
  booktitle={Proceedings of the IEEE/CVF conference on computer vision and pattern recognition},
  pages={20993--21002},
  year={2022}
}

@inproceedings{cui2023sportsmot,
  title={Sportsmot: A large multi-object tracking dataset in multiple sports scenes},
  author={Cui, Yutao and Zeng, Chenkai and Zhao, Xiaoyu and Yang, Yichun and Wu, Gangshan and Wang, Limin},
  booktitle={Proceedings of the IEEE/CVF international conference on computer vision},
  pages={9921--9931},
  year={2023}
}

@inproceedings{cao2023observation,
  title={Observation-centric sort: Rethinking sort for robust multi-object tracking},
  author={Cao, Jinkun and Pang, Jiangmiao and Weng, Xinshuo and Khirodkar, Rawal and Kitani, Kris},
  booktitle={Proceedings of the IEEE/CVF conference on computer vision and pattern recognition},
  pages={9686--9696},
  year={2023}
}

@inproceedings{yang2023hard,
  title={Hard to track objects with irregular motions and similar appearances? make it easier by buffering the matching space},
  author={Yang, Fan and Odashima, Shigeyuki and Masui, Shoichi and Jiang, Shan},
  booktitle={Proceedings of the IEEE/CVF winter conference on applications of computer vision},
  pages={4799--4808},
  year={2023}
}

@inproceedings{yang2024hybrid,
  title={Hybrid-sort: Weak cues matter for online multi-object tracking},
  author={Yang, Mingzhan and Han, Guangxin and Yan, Bin and Zhang, Wenhua and Qi, Jinqing and Lu, Huchuan and Wang, Dong},
  booktitle={Proceedings of the AAAI conference on artificial intelligence},
  volume={38},
  number={7},
  pages={6504--6512},
  year={2024}
}

@article{hu2024trackssm,
  title={Trackssm: A general motion predictor by state-space model},
  author={Hu, Bin and Luo, Run and Liu, Zelin and Wang, Cheng and Liu, Wenyu},
  journal={arXiv preprint arXiv:2409.00487},
  year={2024}
}

@inproceedings{lv2024diffmot,
  title={Diffmot: A real-time diffusion-based multiple object tracker with non-linear prediction},
  author={Lv, Weiyi and Huang, Yuhang and Zhang, Ning and Lin, Ruei-Sung and Han, Mei and Zeng, Dan},
  booktitle={Proceedings of the IEEE/CVF conference on computer vision and pattern recognition},
  pages={19321--19330},
  year={2024}
}

@inproceedings{gao2025multiple,
  title={Multiple object tracking as id prediction},
  author={Gao, Ruopeng and Qi, Ji and Wang, Limin},
  booktitle={Proceedings of the Computer Vision and Pattern Recognition Conference},
  pages={27883--27893},
  year={2025}
}

@inproceedings{hampali2021monte,
  title={Monte carlo scene search for 3d scene understanding},
  author={Hampali, Shreyas and Stekovic, Sinisa and Sarkar, Sayan Deb and Kumar, Chetan S and Fraundorfer, Friedrich and Lepetit, Vincent},
  booktitle={Proceedings of the IEEE/CVF Conference on Computer Vision and Pattern Recognition},
  pages={13804--13813},
  year={2021}
}

@inproceedings{bewley2016simple,
  title={Simple online and realtime tracking},
  author={Bewley, Alex and Ge, Zongyuan and Ott, Lionel and Ramos, Fabio and Upcroft, Ben},
  booktitle={2016 IEEE international conference on image processing (ICIP)},
  pages={3464--3468},
  year={2016},
  organization={Ieee}
}

@inproceedings{wojke2017simple,
  title={Simple online and realtime tracking with a deep association metric},
  author={Wojke, Nicolai and Bewley, Alex and Paulus, Dietrich},
  booktitle={2017 IEEE international conference on image processing (ICIP)},
  pages={3645--3649},
  year={2017},
  organization={IEEE}
}

@article{yu2022relationtrack,
  title={Relationtrack: Relation-aware multiple object tracking with decoupled representation},
  author={Yu, En and Li, Zhuoling and Han, Shoudong and Wang, Hongwei},
  journal={IEEE Transactions on Multimedia},
  volume={25},
  pages={2686--2697},
  year={2022},
  publisher={IEEE}
}

@article{du2023strongsort,
  title={Strongsort: Make deepsort great again},
  author={Du, Yunhao and Zhao, Zhicheng and Song, Yang and Zhao, Yanyun and Su, Fei and Gong, Tao and Meng, Hongying},
  journal={IEEE Transactions on Multimedia},
  volume={25},
  pages={8725--8737},
  year={2023},
  publisher={IEEE}
}

@article{luiten2021hota,
  title={Hota: A higher order metric for evaluating multi-object tracking},
  author={Luiten, Jonathon and Osep, Aljosa and Dendorfer, Patrick and Torr, Philip and Geiger, Andreas and Leal-Taix{\'e}, Laura and Leibe, Bastian},
  journal={International journal of computer vision},
  volume={129},
  number={2},
  pages={548--578},
  year={2021},
  publisher={Springer}
}

@article{ge2021yolox,
  title={Yolox: Exceeding yolo series in 2021},
  author={Ge, Zheng and Liu, Songtao and Wang, Feng and Li, Zeming and Sun, Jian},
  journal={arXiv preprint arXiv:2107.08430},
  year={2021}
}

@article{zhang2021fairmot,
  title={Fairmot: On the fairness of detection and re-identification in multiple object tracking},
  author={Zhang, Yifu and Wang, Chunyu and Wang, Xinggang and Zeng, Wenjun and Liu, Wenyu},
  journal={International journal of computer vision},
  volume={129},
  number={11},
  pages={3069--3087},
  year={2021},
  publisher={Springer}
}

@article{graves2012long,
  title={Long short-term memory},
  author={Graves, Alex},
  journal={Supervised sequence labelling with recurrent neural networks},
  pages={37--45},
  year={2012},
  publisher={Springer}
}

\clearpage

\renewcommand{\theequation}{S\arabic{equation}}
\renewcommand{\thefigure}{S\arabic{figure}}
\renewcommand{\thetable}{S\arabic{table}}
\setcounter{equation}{0}
\setcounter{figure}{0}
\setcounter{table}{0}

% 제목 추가
\title{PlugTrack: Multi-Perceptive Motion Analysis for Adaptive Fusion \\ in Multi-Object Tracking\\[0.5em]
\Large{Supplementary Material}}

\maketitle

\section{Overview}

We provide additional technical details and experimental analyses organized as follows:

\noindent\textbf{Sections A-B:} Mathematical formulations of the Kalman filter and uncertainty quantification method used in our framework.

\noindent\textbf{Section C:} Detailed network architectures of CME components and ABG with layer-wise specifications.

\noindent\textbf{Sections D-F:} Additional experimental results including ablation studies, runtime analysis, and ID switch reduction analysis.

\noindent\textbf{Section G:} Extended qualitative results with frame-by-frame tracking comparisons on DanceTrack sequences.

\noindent\textbf{Section H:} Discussion of future research directions for extending PlugTrack to multiple motion prediction paradigms.

\subsection{A. Kalman Filter Formulation}

The Kalman filter maintains an 8-dimensional state vector for each tracked object:

$$\mathbf{x}_t = [x_t, y_t, w_t, h_t, v_x, v_y, v_w, v_h]^T$$
where $(x_t, y_t, w_t, h_t)$ represent the bounding box center coordinates and dimensions, and $(v_x, v_y, v_w, v_h)$ represent their respective velocities.

The Kalman filter operates under the following linear dynamical system:
\begin{align}
\mathbf{x}_t &= \mathbf{F}\mathbf{x}_{t-1} + \mathbf{w}_t \\
\mathbf{z}_t &= \mathbf{H}\mathbf{x}_t + \mathbf{v}_t
\end{align}
where $\mathbf{F} \in \mathbb{R}^{8 \times 8}$ is the state transition matrix, $\mathbf{H} \in \mathbb{R}^{4 \times 8}$ is the observation matrix, $\mathbf{w}_t \sim \mathcal{N}(0, \mathbf{Q})$ is the process noise, and $\mathbf{v}_t \sim \mathcal{N}(0, \mathbf{R})$ is the measurement noise.

The state transition matrix assumes constant velocity motion:

$$\mathbf{F} = \begin{bmatrix}
\mathbf{I}_4 & \mathbf{I}_4 \\
\mathbf{0}_4 & \mathbf{I}_4
\end{bmatrix}$$

The observation matrix extracts only position measurements:

$$\mathbf{H} = \begin{bmatrix}
\mathbf{I}_4 & \mathbf{0}_4
\end{bmatrix}$$

The Kalman filter recursion consists of prediction and update steps. In the prediction step:
\begin{align}
\hat{\mathbf{x}}_{t|t-1} &= \mathbf{F}\hat{\mathbf{x}}_{t-1|t-1} \\
\mathbf{P}_{t|t-1} &= \mathbf{F}\mathbf{P}_{t-1|t-1}\mathbf{F}^T + \mathbf{Q}
\end{align}

In the update step:
\begin{align}
\mathbf{y}_t &= \mathbf{z}_t - \mathbf{H}\hat{\mathbf{x}}_{t|t-1} \\
\mathbf{S}_t &= \mathbf{H}\mathbf{P}_{t|t-1}\mathbf{H}^T + \mathbf{R} \\
\mathbf{K}_t &= \mathbf{P}_{t|t-1}\mathbf{H}^T\mathbf{S}_t^{-1} \\
\hat{\mathbf{x}}_{t|t} &= \hat{\mathbf{x}}_{t|t-1} + \mathbf{K}_t\mathbf{y}_t \\
\mathbf{P}_{t|t} &= (\mathbf{I}_8 - \mathbf{K}_t\mathbf{H})\mathbf{P}_{t|t-1}
\end{align}
where $\mathbf{y}_t$ is the innovation, $\mathbf{S}_t$ is the innovation covariance, and $\mathbf{K}_t$ is the Kalman gain.

\subsection{B. Uncertainty Quantification via Normalized Innovation Squared}

The innovation $\mathbf{y}_t$ represents the prediction error and follows a Gaussian distribution $\mathbf{y}_t \sim \mathcal{N}(0, \mathbf{S}_t)$. The Normalized Innovation Squared (NIS) is defined as:

$$\epsilon_t = \mathbf{y}_t^T \mathbf{S}_t^{-1} \mathbf{y}_t$$

Under the assumption that the Kalman filter model is correct, $\epsilon_t$ follows a chi-squared distribution with degrees of freedom equal to the dimension of the measurement space (4 in our case): $\epsilon_t \sim \chi^2_4$.

For computational efficiency and dimension-specific uncertainty estimation, we compute the NIS for each dimension independently:

$$\text{NIS}_{t,i} = \frac{y_{t,i}^2}{S_{t,ii}} \quad \text{for } i \in \{x, y, w, h\}$$
where $y_{t,i}$ is the $i$-th component of the innovation vector and $S_{t,ii}$ is the $i$-th diagonal element of the innovation covariance matrix.

To obtain a robust uncertainty measure, we employ a sliding window approach over the most recent $w$ frames (we use $w=3$ in our implementation). For each dimension $i$, we compute:
\begin{align}
\mu_i &= \frac{1}{w} \sum_{j=t-w+1}^{t} \text{NIS}_{j,i} \\
\sigma_i^2 &= \frac{1}{w} \sum_{j=t-w+1}^{t} (\text{NIS}_{j,i} - \mu_i)^2
\end{align}

The final uncertainty measure for dimension $i$ is:

$$\boldsymbol{\sigma}_{\text{KF},i} = \mu_i + \sigma_i$$

This formulation captures both the magnitude of prediction errors through the mean NIS and the consistency of predictions through the standard deviation. The uncertainty vector $\boldsymbol{\sigma}_{\text{KF}} = [\boldsymbol{\sigma}_{\text{KF},x}, \boldsymbol{\sigma}_{\text{KF},y}, \boldsymbol{\sigma}_{\text{KF},w}, \boldsymbol{\sigma}_{\text{KF},h}]^T$ provides dimension-specific confidence measures. Values close to 1 indicate accurate linear motion modeling, while values significantly greater than 1 suggest violation of the linear motion assumption. This uncertainty quantification enables our ABG to make informed decisions about when to trust the Kalman filter predictions versus the data-driven motion predictor for each coordinate independently.

\begin{figure}[h!]
    \centering
    \includegraphics[width=\columnwidth]{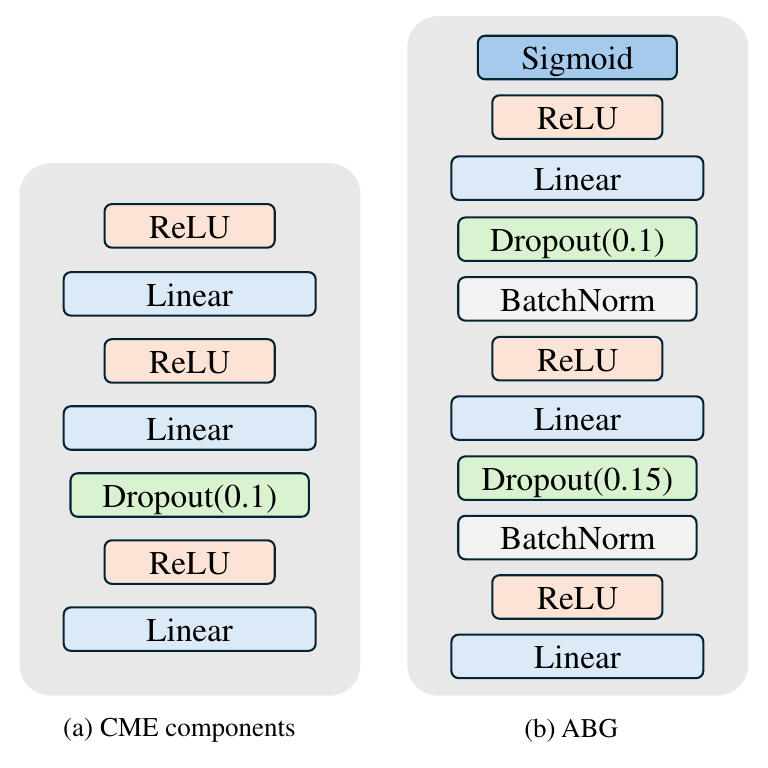} 
    \caption{Network architectures of (a) CME components (PDM, UQM, and Fusion Encoder) and (b) ABG.}
    \label{fig:fig1}
\end{figure}

\subsection{C. Network Architecture Details}
The PDM, UQM, and Fusion Encoder are illustrated in Figure~\ref{fig:fig1}(a). PDM and UQM share identical architectures with three fully connected layers: $\text{Linear}(4\rightarrow64)\rightarrow\text{ReLU}\rightarrow\text{Dropout}(0.1)\rightarrow\text{Linear}(64\rightarrow128)\rightarrow\text{ReLU}\rightarrow\text{Linear}(128\rightarrow32)\rightarrow\text{ReLU}$, mapping 4-dimensional inputs to 32-dimensional feature representations. The Fusion Encoder combines features from the MPM (128-dim), PDM (32-dim), and UQM (32-dim) into a 192-dimensional vector and processes it through $\text{Linear}(192\rightarrow256)\rightarrow\text{ReLU}\rightarrow\text{Dropout}(0.1)\rightarrow\text{Linear}(256\rightarrow512)\rightarrow\text{ReLU}\rightarrow\text{Linear}(512\rightarrow128)\rightarrow\text{ReLU}$, producing the final 128-dimensional multi-perceptive motion feature.

The ABG in Figure~\ref{tab:table1}(b) processes the multi-perceptive motion feature through an architecture with five layers: $\text{Linear}(128\rightarrow256)\rightarrow\text{ReLU}\rightarrow\text{BatchNorm}\rightarrow\text{Dropout}(0.15)\rightarrow\text{Linear}(256\rightarrow128)\rightarrow\text{ReLU}\rightarrow\text{BatchNorm}\rightarrow\text{Dropout}(0.1)\rightarrow\text{Linear}(128\rightarrow4)\rightarrow\text{ReLU}\rightarrow\text{Sigmoid}$. The network incorporates batch normalization and dropout for regularization to generate optimal alpha blending factors.

\begin{table}[h!]
\centering
\small
\begin{tabular}{c|c|c|ccc}
\hline
MPM & PDM & UQM & HOTA & AssA  & IDF1  \\
\hline
\multicolumn{6}{c}{\textbf{Ours (TrackSSM)}} \\
\hline
           &            &            & 53.4 & 36.3 & 53.3 \\
\checkmark &            &            & 54.5 & 37.7 & 54.7 \\
\checkmark & \checkmark &            & 54.5 & 37.5 & 54.8 \\
\checkmark &            & \checkmark & 54.6 & 37.8 & 54.7 \\
\checkmark & \checkmark & \checkmark & \textbf{54.9} & \textbf{38.1} & \textbf{54.9} \\
\hline
\end{tabular}
\caption{Ablation study of CME Analyzers on DanceTrack validation set. The best results are shown in bold.}
\label{tab:table1}
\end{table}

\subsection{D. Ablation Study on CME Components}

To understand each component's contribution to multi-perceptive analysis, we conduct systematic ablations. Table~\ref{tab:table1} analyzes the contribution of each CME module on DanceTrack validation set. The baseline configuration (first row) directly feeds the tracklet to ABG without specialized modules, achieving 53.4 HOTA. Adding only Motion Pattern Module (MPM) improves HOTA by 1.1 to 54.5, demonstrating temporal pattern analysis value. Combining MPM with PDM maintains performance at 54.5 HOTA, while MPM with UQM achieves 54.6 HOTA, showing that uncertainty information provides complementary benefits. The full configuration with all three modules achieves the best performance of 54.9 HOTA, with AssA improving from 36.3 to 38.1 and IDF1 from 53.3 to 54.9. This demonstrates that multi-perceptive motion analysis creates synergistic interactions: MPM captures temporal patterns, PDM models prediction discrepancies, and UQM quantifies uncertainty, with their combination enabling more robust and accurate tracking performance.

\begin{table}[h!]
\centering
\small
\begin{tabular}{c|cc|cc}
\hline
\multirow{2}{*}{Detector} & \multicolumn{2}{c|}{TrackSSM} & \multicolumn{2}{c}{DiffMOT} \\
& Base & Ours & Base & Ours \\
\hline
YOLOX-x & 37.2 & 34.2 & 25.9 & 24.7 \\
YOLOX-l & 39.3 & 36.4 & 27.3 & 25.8 \\
YOLOX-m & 43.3 & 40.8 & 29.6 & 27.2 \\
YOLOX-s & 47.9 & 45.0 & 30.8 & 28.1 \\
\hline
\end{tabular}
\caption{FPS comparison between baseline trackers and our method with different YOLOX detector sizes.}
\label{tab:table2}
\end{table}

\subsection{E. Runtime Analysis with Different Detectors}

Table~\ref{tab:table2} presents the runtime performance comparison between baseline trackers (TrackSSM and DiffMOT) and PlugTrack across different YOLOX detector sizes. For TrackSSM, our method introduces a modest computational overhead, reducing FPS from 37.2 to 34.2 (-8.1\%) with YOLOX-x, and from 47.9 to 45.0 (-6.1\%) with YOLOX-s, with the relative overhead decreasing as the detector becomes lighter, indicating good scalability. Similarly, for DiffMOT, our approach reduces FPS from 25.9 to 24.7 (-4.6\%) with YOLOX-x and from 30.8 to 28.1 (-8.8\%) with YOLOX-s. Notably, both our enhanced TrackSSM and DiffMOT maintain real-time performance across all detector configurations, demonstrating that the computational cost of our CME and ABG is relatively constant regardless of detector size, resulting in a smaller relative impact on faster configurations. This shows that our method can be effectively integrated into existing data-driven motion predictors without compromising their real-time capabilities, making it practical for deployment in time-critical applications.

\begin{table}[h!]
\centering
\small
\begin{tabular}{c|cc|cc}
\hline
\multirow{2}{*}{Dataset} & \multicolumn{2}{c|}{TrackSSM} & \multicolumn{2}{c}{DiffMOT} \\
& Base & Ours & Base & Ours \\
\hline
DanceTrack (val) & 1961 & 1861 & 1970 & 1839 \\
MOT17 (test) & 3531 & 3153 & 2583 & 2361 \\
MOT20 (test) & 2015 & 1526 & 1665 & 1549 \\
\hline
\end{tabular}
\caption{ID switch (IDSW) comparison between baseline trackers and our method across different datasets. Lower values indicate better performance.}
\label{tab:table3}
\end{table}

\subsection{F. Analysis of ID Switch Reduction}

Table~\ref{tab:table3} demonstrates the effectiveness of our method in reducing ID switches (IDSW) across different datasets. ID switches occur when the tracker incorrectly assigns a detection to a different identity, which is a critical failure mode in MOT. Our method consistently reduces IDSW across all datasets and both baseline predictors. For TrackSSM, we achieve reductions on DanceTrack (1961→1861, -5.1\%), MOT17 (3531→3153, -10.7\%), and MOT20 (2015→1526, -24.3\%). Similarly, for DiffMOT, our approach reduces IDSW on DanceTrack (1970→1839, -6.6\%), MOT17 (2583→2361, -8.6\%), and MOT20 (1665→1549, -7.0\%). The most significant improvement is observed on MOT20 with TrackSSM, where our method reduces ID switches by nearly a quarter. This substantial reduction in ID switches can be attributed to our adaptive blending mechanism, which intelligently combines Kalman filter predictions with learned motion patterns based on contextual information. By leveraging uncertainty quantification and prediction discrepancy analysis, our method makes more informed decisions during challenging scenarios such as occlusions, rapid motion changes, and crowded scenes, thereby maintaining more consistent object identities throughout the tracking sequence.

\subsection{G. Qualitative Results}

Figure~\ref{fig:fig2} presents qualitative comparisons between our PlugTrack method and baseline data-driven motion predictors on challenging sequences from the DanceTrack dataset. Each sequence shows tracking results with bounding box colors indicating different trackers: yellow (TrackSSM), green (DiffMOT), orange (PlugTrack(TrackSSM)), blue (PlugTrack(DiffMOT)), and red (ground truth). The IoU scores between predictions and ground truth are displayed in the top-left corner of each frame. In the top row (DanceTrack0004), PlugTrack maintains superior tracking accuracy throughout the sequence, particularly during complex dance movements between frames 135-138. At frame 136, baseline TrackSSM and DiffMOT achieve IoU scores of 37.16 and 37.22 respectively, but when integrated with PlugTrack, their performance significantly improves to 60.23 and 55.12 respectively. The middle row (DanceTrack0019) demonstrates PlugTrack's robust handling of rapid pose changes and partial occlusions during frames 427-430, where dancers perform floor movements with significant deformation. PlugTrack achieves substantially higher IoU scores, with PlugTrack(TrackSSM) reaching 74.42 compared to baseline TrackSSM's 63.25 at frame 429. The bottom row (DanceTrack0063) showcases a particularly challenging scenario with multiple overlapping dancers and dramatic lighting effects. PlugTrack maintains more stable tracking performance, with the most significant improvement at frame 68 where PlugTrack(TrackSSM) achieves 40.44 IoU compared to baseline TrackSSM's 14.38. These qualitative results confirm that our adaptive blending strategy effectively combines the complementary strengths of Kalman filter and data-driven motion predictor.

\subsection{H. Future Work}

While PlugTrack currently demonstrates successful integration with data-driven motion predictors, our framework's adaptive fusion principle can be extended to accommodate a broader range of TBD paradigms. A promising future direction involves extending PlugTrack to incorporate Kalman filter-based heuristic methods such as OC-SORT, C-BIoU, and Hybrid-SORT. By treating these enhanced Kalman variants as additional motion predictors within our framework, we could enable adaptive fusion among multiple Kalman-based approaches. For instance, we could leverage OC-SORT's observation-centric recovery during occlusions and C-BIoU's buffered matching in crowded scenes. Our CME could learn to identify scenarios where each heuristic modification provides optimal predictions. Meanwhile, the ABG could dynamically blend predictions from multiple Kalman variants based on the current motion context. This extended framework could potentially support simultaneous fusion of Kalman filters, heuristic-enhanced variants, and data-driven predictors within a unified system. Such an approach would require enhanced multi-perceptive analysis and multi-way blending factors to leverage the collective strengths of diverse prediction methodologies.

\begin{figure*}[t]
    \centering
    \includegraphics[width=\textwidth]{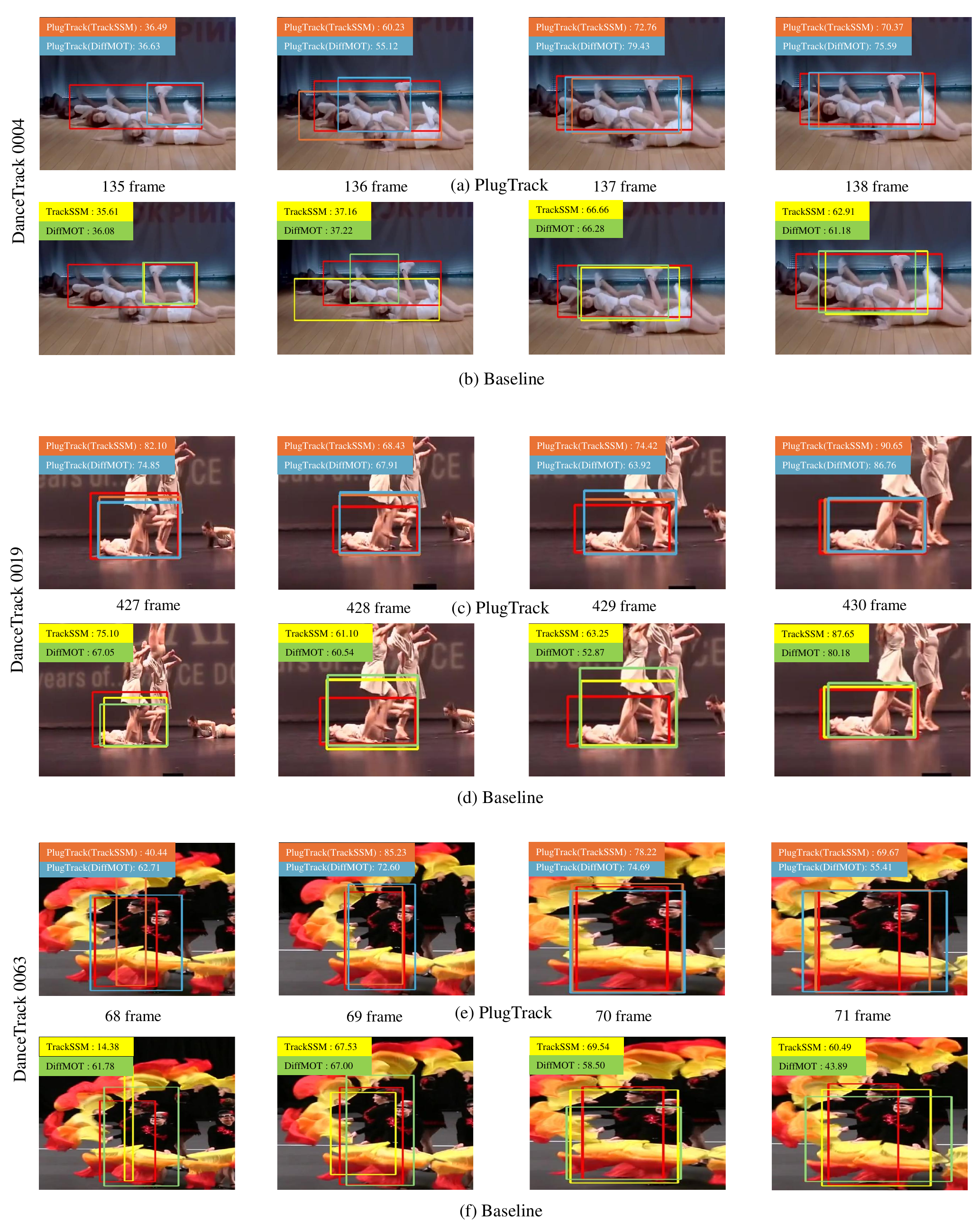} 
    \caption{Qualitative comparison on DanceTrack sequences. Each row shows tracking results from (a), (c), (e) PlugTrack and (b), (d), (f) baseline methods. Bounding box colors indicate different predictors: yellow (TrackSSM), green (DiffMOT), orange (PlugTrack(TrackSSM)), blue (PlugTrack(DiffMOT)), and red (ground truth). The values in the top-left corner represent the IoU between predictions and ground truth. PlugTrack consistently achieves higher IoU scores across all sequences, demonstrating improved tracking accuracy.}
    \label{fig:fig2}
\end{figure*}

\end{document}